\documentclass[11pt]{article}

\usepackage[final]{acl}

\usepackage{times}
\usepackage{latexsym}
\usepackage[T1]{fontenc}
\usepackage[utf8]{inputenc}
\usepackage{microtype}
\usepackage{inconsolata}
\usepackage{graphicx}
\usepackage{caption}

\usepackage{subfigure}
\usepackage{booktabs}
\usepackage{adjustbox}
\usepackage{multirow}
\usepackage{makecell}
\usepackage{float}
\usepackage{placeins}
\usepackage{cuted} % strip: full-width blocks in twocolumn without a separate float page
\usepackage{xspace}
\usepackage{xcolor}
\usepackage{enumitem}

\usepackage{amsmath}
\usepackage{amssymb}
\usepackage{mathtools}
\usepackage{amsthm}
\usepackage[capitalize,noabbrev]{cleveref}

\usepackage{algorithm}
\usepackage{algorithmic}

\usepackage{soul}
\usepackage{listings}
\usepackage{tikz}
\usepackage{colortbl}
\usepackage{setspace}

\usepackage{wasysym} % provides \Letter
\usepackage{orcidlink}
\usepackage{xurl} % better link breaking

\setlist[itemize]{noitemsep, topsep=0pt, leftmargin=*}
\setlist[enumerate]{noitemsep, topsep=0pt, leftmargin=*}

\newcommand{\retriever}{\textsc{Adaptive Retriever of Knowledge}\xspace}
\newcommand{\retrieverShort}{\textsc{ARK}\xspace}
\newcommand{\retrieverNoSc}{Adaptive Retriever of Knowledge\xspace}

\newcommand{\slimparagraph}[1]{\noindent{\textbf{#1}}}

\setstcolor{orange}

\definecolor{fuchsia}{RGB}{255,0,255}
\setstcolor{fuchsia}

\newcommand{\unbreakable}[1]{
  \relpenalty=10000%
  \binoppenalty=10000%
  #1%
}

\DeclareMathOperator{\rel}{rel}

\makeatletter
\newcommand{\tbdcite}{%
  \@ifnextchar\bgroup{\@tbdciteWith}{\@tbdciteNoArg}%
}
\newcommand{\@tbdciteNoArg}{%
  {\color{brown}[\@tbd@repeat{1}]}\xspace%
}
\newcommand{\@tbdciteWith}[1]{%
  {\color{brown}[\@tbd@repeat{#1}]}\xspace%
}
\newcommand{\@tbd@repeat}[1]{%
  \@tempcnta=0\relax%
  \@whilenum\@tempcnta<#1\do{%
    cite%
    \advance\@tempcnta\@ne%
    \ifnum\@tempcnta<#1 ,\space\fi%
  }%
}
\makeatother

\newcommand{\FigMain}[1][t]{
    \begin{figure*}[#1]
        \vskip 0.2in
        \begin{center}
            \includegraphics[width=1\textwidth]{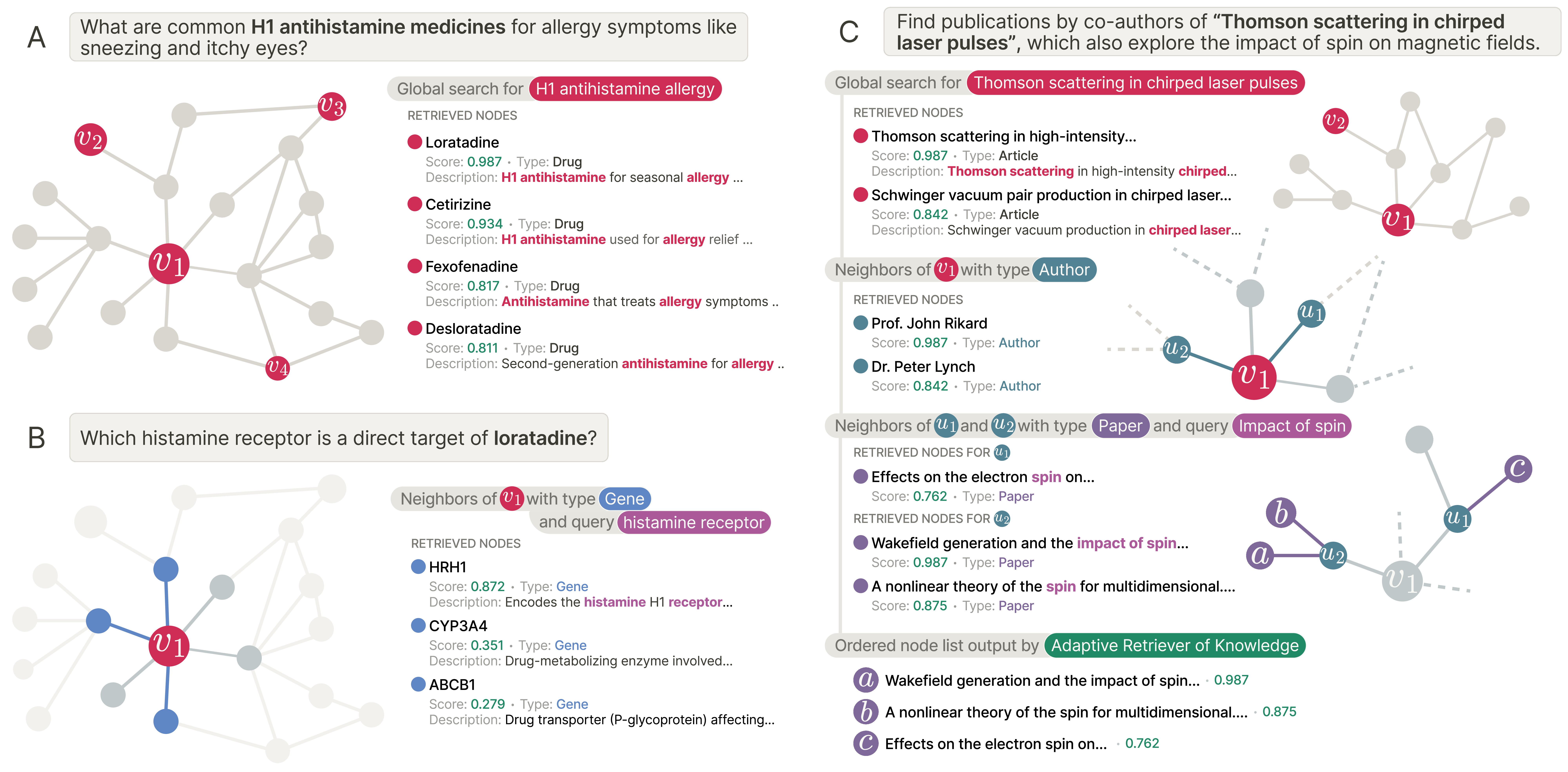}
            \caption{
            Overview of \retriever.
            \retrieverShort interacts with a KG through a minimal two-tool interface:
            \textbf{(a)} For text-dominant queries, \retrieverShort emphasizes breadth by issuing \textsc{Global Search} to retrieve a broad set of candidates.
            \textbf{(b)} For relation-focused queries, \retrieverShort applies \textsc{Neighborhood Exploration} starting from a previously retrieved node (in this case, a drug) and expanding to related entities, enabling targeted relational retrieval.
            \textbf{(c)} For relation-dominant queries, \retrieverShort performs multi-hop retrieval by alternating \textsc{Global Search} and \textsc{Neighborhood Exploration}: it retrieves an initial node (\textit{e.g.}, an article), expands to related entities (\textit{e.g.}, co-authors), and continues expanding and filtering (\textit{e.g.}, papers connected to each author that match query keywords) to recover an ordered set of relevant evidence.
            }
            \label{fig:main-figure}
        \end{center}
        \vskip -0.1in
    \end{figure*}
}

\newcommand{\FigAdaptiveToolUse}[1][t]{
    \begin{figure}[#1]
        \centering
        \includegraphics[width=0.48\textwidth]{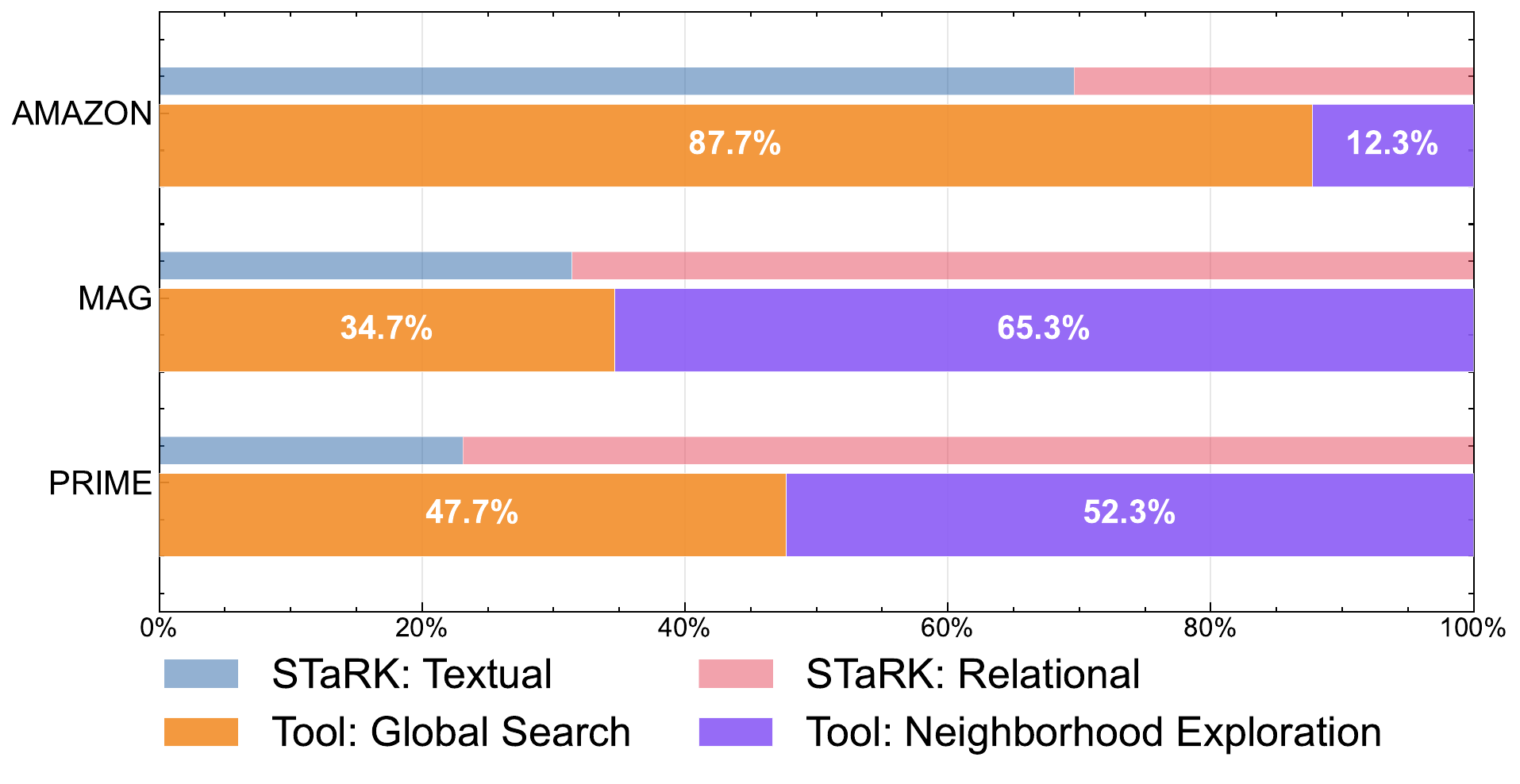}
        \caption{
        Thin bars show the share of text- vs.~relation-centric queries in STaRK; thick bars show \retrieverShort's tool-call use. These STaRK annotations are not provided to \retrieverShort; instead, \retrieverShort autonomously shifts tool use to match the dominant query type.
        }
        \label{fig:adaptive-tool-use}
        \vskip -0.2in
    \end{figure}
}

\newcommand{\FigNAgents}[1][t]{
    \begin{figure*}[#1]
        \vskip 0.2in
        \begin{center}
            \includegraphics[width=1\textwidth]{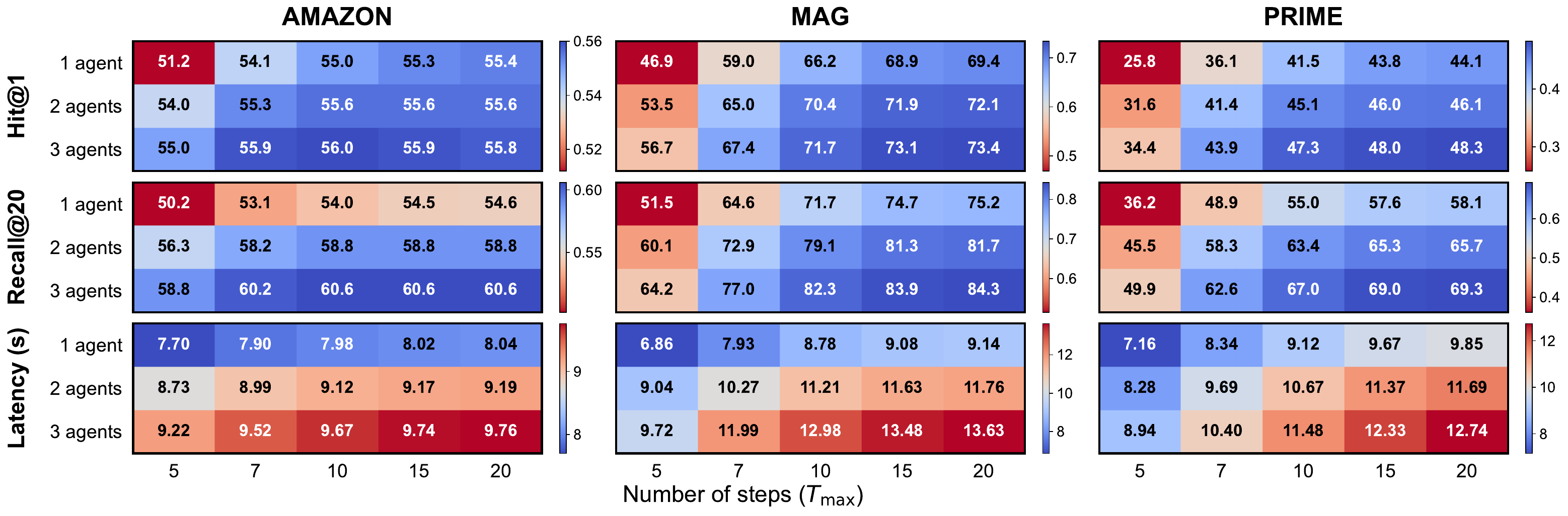}
            \caption{Retrieval quality and latency as a function of inference-time budget. Heatmaps report Hit@1, Recall@20, and end-to-end latency (seconds) on each STaRK graph. Moving from the top left (shallow trajectories, single agent) to the bottom right (deeper trajectories, multi-agent) allocates more compute and improves retrieval performance at the cost of higher latency. Color scales are normalized within each graph and metric for readability.}
            \label{fig:n-agents}
        \end{center}
        \vskip -0.2in
    \end{figure*}
}

\newcommand{\FigSFTImpact}[1][t]{
    \begin{figure*}[#1]
        \begin{center}
            \includegraphics[width=\textwidth]{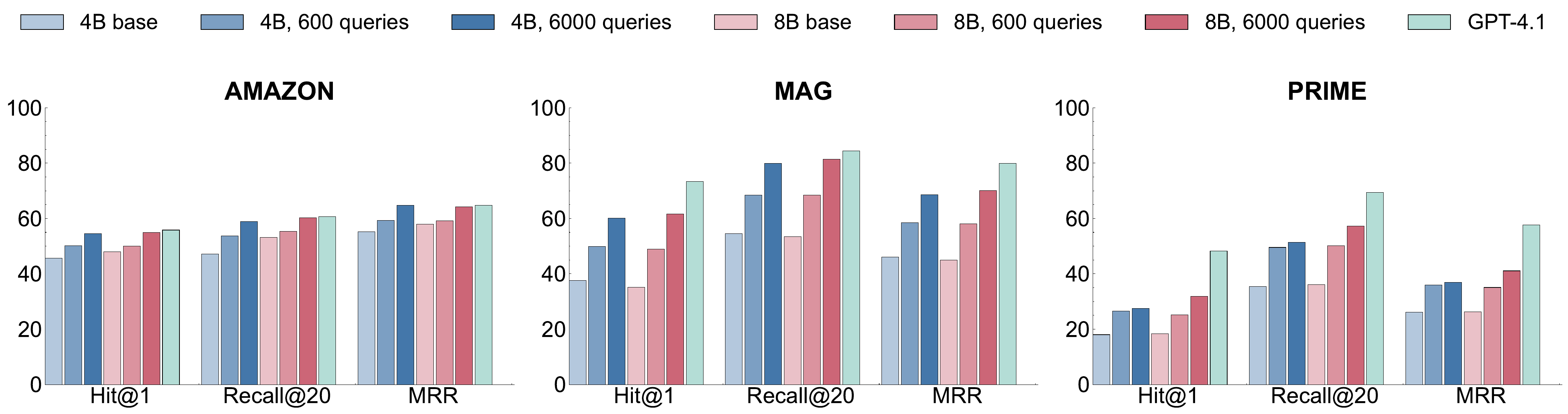}%
            \caption{Evaluation of the same \retrieverShort pipeline on the STaRK test sets while varying the underlying model and the distillation budget. ``600 queries'' and ``6000 queries'' denote Qwen backbones fine-tuned on trajectories generated by GPT-4.1 from 600 or 6000 training queries per graph, respectively (three trajectories per query; tool calls and observations only; no label supervision).}
            \label{fig:sft-impact}
        \end{center}
    \end{figure*}
}

\newcommand{\FigNeighborExploreDistribution}[1][t]{%
    \begin{figure}[#1]
        \centering
        \includegraphics[width=0.48\textwidth]{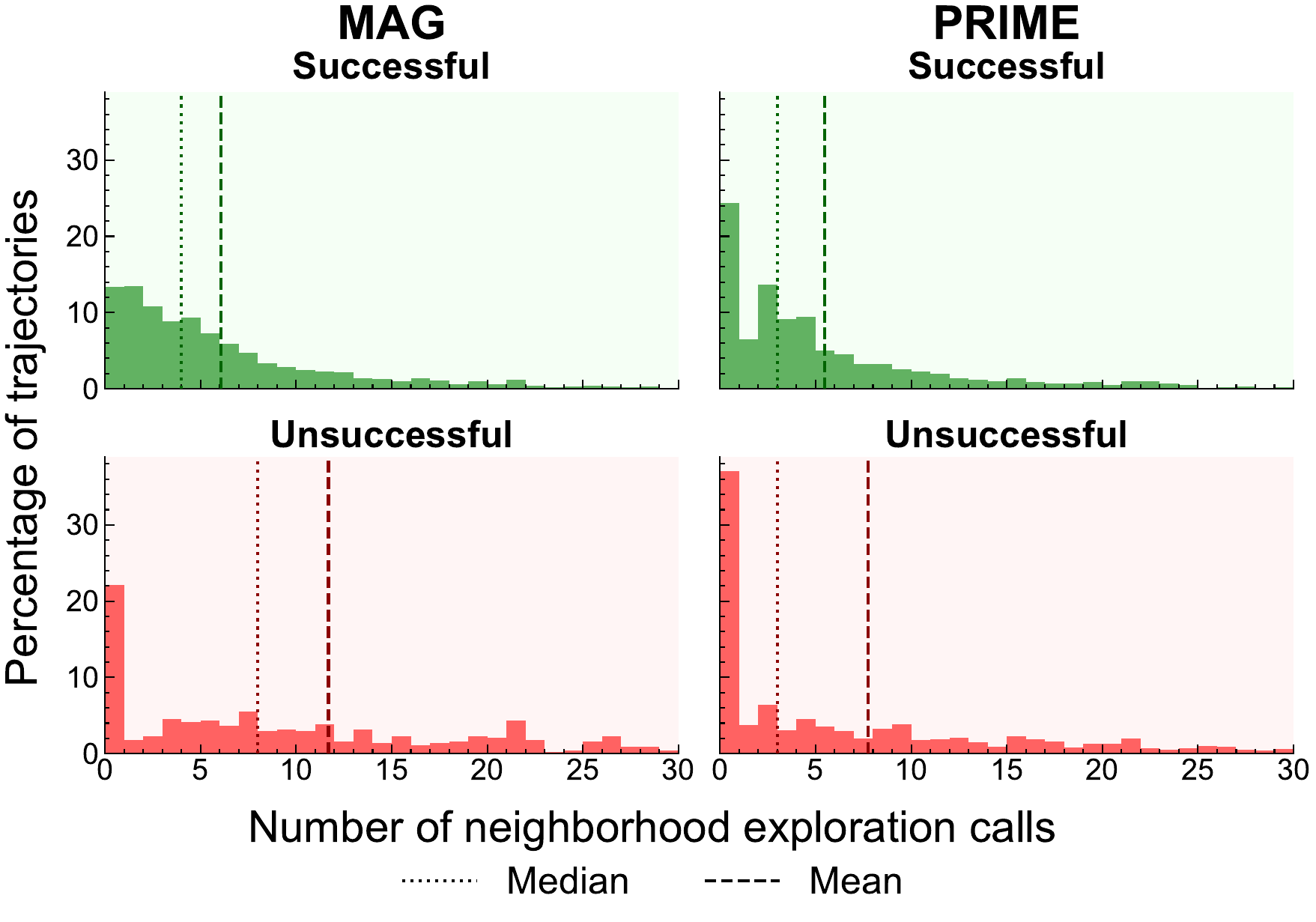}
        \caption{Distribution of the number of neighborhood exploration calls, split by successful (Hit@5) and unsuccessful trajectories.}
        \label{fig:neighbor-explore-distribution}
        \vskip -0.1in
    \end{figure}
}

\definecolor{tabledarkgreen}{RGB}{116,198,157}
\definecolor{tablelightgreen}{RGB}{216,243,220}
\definecolor{tabledarkblue}{RGB}{112,186,255}
\definecolor{tablelightblue}{RGB}{222,239,254}

\newcommand{\bestoverall}[1]{\textbf{#1}}

\newcommand{\besttf}[1]{\cellcolor{tabledarkgreen}{#1}}
\newcommand{\secondtf}[1]{\cellcolor{tablelightgreen}{#1}}

\newcommand{\besttr}[1]{\cellcolor{tabledarkblue}{#1}}
\newcommand{\secondtr}[1]{\cellcolor{tablelightblue}{#1}}

\newcommand\mybox[2][]{
  \hspace{1.5pt}\tikz[overlay]\node[fill=blue!20, inner xsep=2pt, inner ysep=0pt, outer xsep=10pt, anchor=text, rectangle, rounded corners=1mm, #1] {#2};\phantom{#2}\hspace{1.5pt}
}

\newcommand{\TabStarkResults}[1][!ht]{
    \begin{table*}[!ht]
        \centering
        \resizebox{\textwidth}{!}{
            \begin{tabular}{llcccccccccccc}
                \toprule
                         &                                               & \multicolumn{3}{c}{AMAZON} & \multicolumn{3}{c}{MAG} & \multicolumn{3}{c}{PRIME} & \multicolumn{3}{c}{Average} \\
                Category & Method
                         & Hit@1                                         & R@20                       & MRR
                         & Hit@1                                         & R@20                       & MRR
                         & Hit@1                                         & R@20                       & MRR
                         & Hit@1                                         & R@20                       & MRR \\
                \midrule

                \multirow{9}{*}{\makecell[l]{\textit{Training-free}}}
                         & \multicolumn{13}{l}{\textit{Retrieval-based}} \\
                         & BM25                                          & 44.94                      & 53.77                   & 55.30                     & 25.85                       & 45.69             & 34.91             & 12.75             & 31.25             & 19.84             & 27.85             & 43.57             & 36.68             \\
                         & ada-002                                       & 39.16                      & 53.29                   & 50.35                     & 29.08                       & 48.36             & 38.62             & 12.63             & 36.00             & 21.41             & 26.96             & 45.88             & 36.79             \\
                         & GritLM-7B
& 42.08 & 56.52 & 53.46
& 37.90 & 46.40 & 47.25
& 15.57 & 39.09 & 24.11
& 31.85 & 47.34 & 41.61 \\
                         & KAR                                           & \secondtf{54.20}           & \secondtf{57.24}        & \secondtf{61.29}          & \secondtf{50.47}            & \secondtf{60.28}  & \secondtf{57.51}  & \secondtf{30.35}  & 50.81             & \secondtf{39.22}  & \secondtf{45.01}  & \secondtf{56.11}  & \secondtf{52.67}  \\
                         & \multicolumn{13}{l}{\textit{Agent-based}} \\
                         & Think-on-Graph + GPT-4o                             & 20.67                      & 25.81                   & 30.90                     & 23.33                       & 48.03             & 36.38             & 16.67             & \secondtf{54.35}  & 27.02             & 20.22             & 42.73             & 31.43             \\
                         & Think-on-Graph + LLaMA3                         & 4.21                       & 2.61                    & 5.25                      & 12.00                       & 6.77              & 12.67             & 21.92             & 33.84             & 26.61             & 12.71             & 14.41             & 14.84             \\
                         & \retrieverShort                                  & \besttf{\bestoverall{55.82}} & \besttf{60.61}          & \besttf{\bestoverall{64.77}} & \besttf{\bestoverall{73.40}} & \besttf{\bestoverall{84.47}} & \besttf{\bestoverall{79.87}} & \besttf{48.20}    & \besttf{69.46}    & \besttf{57.68}    & \besttf{\bestoverall{59.14}} & \besttf{\bestoverall{71.51}} & \besttf{\bestoverall{67.44}} \\

                \midrule

                \multirow{5}{*}{\makecell[l]{\textit{Requires training}\\\textit{on target graph}}}
                         & \multicolumn{13}{l}{\textit{Retrieval-based}} \\
                         & mFAR                                          & \secondtr{53.0}           & \besttr{\bestoverall{66.3}}   & \besttr{64.3}             & 55.9                        & 74.1              & 64.3              & \secondtr{40.0}  & \secondtr{72.6}  & \secondtr{52.0}  & \besttr{49.63}    & \besttr{71.00}    & \besttr{60.20}    \\
                         & MoR                                           & 52.19                      & 59.92                   & 62.24                     & \secondtr{58.19}           & \secondtr{75.01} & \secondtr{67.14} & 36.41             & 63.48             & 46.92             & 48.93             & 66.14             & \secondtr{58.77} \\
                         & \multicolumn{13}{l}{\textit{Agent-based}} \\
                         & GraphFlow                                     & 47.85                      & 36.15                   & 55.49                     & 39.09                       & 57.18             & 47.82             & \besttr{\bestoverall{51.39}}    & \besttr{\bestoverall{79.71}}    & \besttr{\bestoverall{61.37}}    & 46.11             & 57.68             & 54.89             \\
                         & AvaTaR                                        & 49.90                      & \secondtr{60.60}       & 58.70                     & 44.36                       & 50.63             & 51.15             & 18.40             & 39.30             & 26.73             & 37.55             & 50.18             & 45.53             \\
                         & \retrieverShort distilled                     & \besttr{54.99}             & 60.31                   & \secondtr{64.24}         & \besttr{61.66}              & \besttr{81.39}    & \besttr{70.09}    & 31.87             & 57.22             & 41.08             & \secondtr{49.51} & \secondtr{66.31} & 58.47             \\

                \bottomrule
            \end{tabular}
        }
        \caption{Retrieval performance on STaRK synthetic test sets. \mybox[fill=tabledarkgreen!100]{\strut Dark green} and \mybox[fill=tablelightgreen!100]{\strut light green} indicate best and second-best in the training-free category, respectively. \mybox[fill=tabledarkblue!100]{\strut Dark blue} and \mybox[fill=tablelightblue!100]{\strut light blue} indicate best and second-best in the requires-training category, respectively. \textbf{Bold} indicates the best result overall for each metric column.}
        \label{tab:stark-results}
    \end{table*}
}

\newcommand{\TabToolsetAblation}[1][!ht]{
    \begin{table}[#1]
        \centering
        \adjustbox{width=\columnwidth,center}{
            \begin{tabular}{lcccccc}
                \toprule
                Setup                    & \multicolumn{2}{c}{AMAZON} & \multicolumn{2}{c}{MAG} & \multicolumn{2}{c}{PRIME}                                                          \\
                \cmidrule(lr){2-3} \cmidrule(lr){4-5} \cmidrule(lr){6-7}
                                         & Hit@1                      & R@20                    & Hit@1                     & R@20             & Hit@1            & R@20             \\
                \midrule
                Full                     & \textbf{58.5}              & \textbf{60.2}           & \textbf{79.2}             & \textbf{83.3}    & \textbf{49.2}    & \textbf{73.3}    \\
                \midrule
                w/o $\operatorname{Neighbors}$ & 54.5                       & 55.4                    & 30.5                      & 39.4             & 23.1             & 40.5             \\
                $\operatorname{Neighbors}$ w/o $q$ & \underline{56.0}           & 57.9                    & 72.1                      & 79.8             & \underline{44.7} & \underline{68.3} \\
                $\operatorname{Neighbors}$ w/o $F$ & 55.5                       & \underline{59.9}        & \underline{79.2}         & \underline{84.8} & 42.2             & 65.0             \\
                \bottomrule
            \end{tabular}
        }
        \caption{Impact of toolset design on retrieval performance across graphs. w/o $\operatorname{Neighbors}$ removes neighborhood exploration entirely. $\operatorname{Neighbors}$ w/o $q$  disables query-based ranking within the neighborhood, and $\operatorname{Neighbors}$ w/o $F$ disables type-based filtering.
        Results are reported on a random 10\% subset of the test split.}
        \label{tab:toolset-ablation}
        \vskip -0.15in
    \end{table}
}

\newcommand{\TabAggregationComparison}[1][!ht]{
    \begin{table}[#1]
        \centering
        \adjustbox{width=\columnwidth,center}{
            \begin{tabular}{lcccccc}
                \toprule
                Setup & \multicolumn{2}{c}{AMAZON} & \multicolumn{2}{c}{MAG} & \multicolumn{2}{c}{PRIME} \\
                \cmidrule(lr){2-3} \cmidrule(lr){4-5} \cmidrule(lr){6-7}
                 & Hit@1 & R@20 & Hit@1 & R@20 & Hit@1 & R@20 \\
                \midrule
                Voting   & \textbf{55.82} & \textbf{60.61} & \textbf{73.40} & \textbf{84.47} & \textbf{48.20} & \textbf{69.46} \\
                Ordering & 55.39 & 60.06 & 71.24 & 84.15 & 44.70 & 68.78 \\
                Random   & 17.65 & 57.55 & 38.04 & 83.03 & 26.30 & 67.50 \\
                \bottomrule
            \end{tabular}
        }
        \caption{Comparison of strategies for aggregating the outputs of three parallel agents. Voting ranks nodes by frequency across trajectories and breaks ties by earliest occurrence. Ordering concatenates trajectories and keeps first appearance, while Random shuffles the union of retrieved nodes.}
        \label{tab:aggregation-comparison}
    \end{table}
}

\newcommand{\TabDenseRetrieverAblation}[1][!ht]{
    \begin{table}[#1]
        \centering
        \adjustbox{width=\columnwidth,center}{
            \begin{tabular}{lcccccc}
                \toprule
                Setup & \multicolumn{2}{c}{AMAZON} & \multicolumn{2}{c}{MAG} & \multicolumn{2}{c}{PRIME} \\
                \cmidrule(lr){2-3} \cmidrule(lr){4-5} \cmidrule(lr){6-7}
                 & Hit@1 & R@20 & Hit@1 & R@20 & Hit@1 & R@20 \\
                \midrule
                BM25  & \textbf{55.82} & \textbf{60.61} & 73.40 & 84.47 & \textbf{48.20} & \textbf{69.46} \\
                Dense & 47.13 & 58.13 & \textbf{75.88} & \textbf{85.11} & 43.23 & 68.58 \\
                \bottomrule
            \end{tabular}
        }
        \caption{BM25 vs.~dense retrieval inside \retrieverShort on the full STaRK synthetic test sets. The dense retriever uses \texttt{text-embedding-3-large}.}
        \label{tab:dense-ablation}
    \end{table}
}

\newcommand{\TabBudgetControlledComparison}[1][!ht]{
    \begin{table}[#1]
        \centering
        \adjustbox{width=\columnwidth,center}{
            \begin{tabular}{lcccccc}
                \toprule
                Setup & \multicolumn{2}{c}{AMAZON} & \multicolumn{2}{c}{MAG} & \multicolumn{2}{c}{PRIME} \\
                \cmidrule(lr){2-3} \cmidrule(lr){4-5} \cmidrule(lr){6-7}
                 & Hit@1 & R@20 & Hit@1 & R@20 & Hit@1 & R@20 \\
                \midrule
                1 agent, $T_{\max}=30$ & 55.4 & 54.6 & 69.5 & 75.4 & 44.3 & 58.3 \\
                3 agents, $T_{\max}=10$ & \textbf{56.0} & \textbf{60.6} & \textbf{71.7} & \textbf{82.3} & \textbf{47.3} & \textbf{67.0} \\
                \bottomrule
            \end{tabular}
        }
        \caption{Cost-controlled comparison between a single longer trajectory and three shorter parallel trajectories under the same total step budget (30 steps).}
        \label{tab:budget-controlled}
    \end{table}
}

\newcommand{\TabGlobalAfterFirstStep}[1][!ht]{
    \begin{table}[#1]
        \centering
        \small
        \begin{tabular}{lcc}
            \toprule
            Dataset & All traj. & Successful traj. \\
            \midrule
            AMAZON & 10.1 & 8.9 \\
            MAG    & 25.2 & 20.5 \\
            PRIME  & 36.9 & 35.8 \\
            \bottomrule
        \end{tabular}
        \caption{Percentage of trajectories that invoke \textsc{Global Search} after step 1. Successful trajectories are those with Hit@5.}
        \label{tab:global-after-first-step}
    \end{table}
}

\newcommand{\TabTrajectoryToolMixing}[1][!ht]{
    \begin{table}[#1]
        \centering
        \small
        \begin{tabular}{llcc}
            \toprule
            Split & Dataset & \makecell{Only global\\search} & Both tools \\
            \midrule
            \multirow{3}{*}{All traj.}
                & AMAZON & 79.4 & 20.6 \\
                & MAG    & 13.3 & 86.7 \\
                & PRIME  & 24.4 & 75.6 \\
            \midrule
            \multirow{3}{*}{Successful traj.}
                & AMAZON & 80.7 & 19.3 \\
                & MAG    & 14.9 & 85.1 \\
                & PRIME  & 29.2 & 70.8 \\
            \bottomrule
        \end{tabular}
        \caption{Trajectory-level tool usage percentages. Since every trajectory begins with \textsc{Global Search}, runs are either global-search-only or use both tools. Successful trajectories are those with Hit@5.}
        \label{tab:trajectory-tool-mixing}
    \end{table}
}

\newcommand{\TabAppendixResults}{%
    \begin{strip}
        \centering
        \resizebox{\textwidth}{!}{%
            \begin{tabular}{lcccccccccccc}
                \toprule
                              & \multicolumn{4}{c}{AMAZON} & \multicolumn{4}{c}{MAG} & \multicolumn{4}{c}{PRIME}                                                                         \\
                Method        & Hit@1                      & Hit@5                   & R@20                      & MRR   & Hit@1 & Hit@5 & R@20  & MRR   & Hit@1 & Hit@5 & R@20  & MRR   \\
                \midrule
                GPT-4o                               & 55.13 & 76.37 & 57.18 & 64.29 & 67.01 & 86.67 & 79.79 & 75.46 & 36.01 & 60.17 & 60.13 & 46.44 \\
                GPT-4.1                              & 55.82 & 75.80 & 60.61 & 64.77 & 73.40 & 87.92 & 84.47 & 79.87 & 48.20 & 69.57 & 69.46 & 57.68 \\
                GPT-5.2                              & 56.12 & 77.76 & 58.62 & 65.71 & 73.13 & 91.11 & 86.21 & 81.01 & 50.64 & 71.44 & 70.04 & 59.55 \\
                Qwen3-4B                             & 45.69 & 67.46 & 47.17 & 55.26 & 37.64 & 56.56 & 54.54 & 46.01 & 18.02 & 37.00 & 35.43 & 26.20 \\
                Qwen3-4B-600                          & 50.10 & 71.26 & 53.73 & 59.32 & 49.90 & 69.21 & 68.47 & 58.43 & 26.52 & 47.27 & 49.54 & 36.01 \\
                Qwen3-4B-6000                         & 54.46 & 74.66 & 58.91 & 64.78 & 60.16 & 78.79 & 79.88 & 68.59 & 27.46 & 49.15 & 51.40 & 36.95 \\
                Qwen3-8B                             & 47.95 & 70.03 & 53.10 & 57.95 & 35.09 & 57.02 & 53.42 & 44.96 & 18.37 & 36.41 & 36.13 & 26.28 \\
                Qwen3-8B-600                          & 50.06 & 70.07 & 55.29 & 59.13 & 48.89 & 69.41 & 68.48 & 58.13 & 25.24 & 47.20 & 50.16 & 35.08 \\
                Qwen3-8B-6000                         & 54.99 & 74.35 & 60.31 & 64.24 & 61.66 & 80.41 & 81.39 & 70.09 & 31.87 & 51.10 & 57.22 & 41.08 \\
                \bottomrule
            \end{tabular}}
        \captionof{table}{Retrieval performance on STaRK synthetic test sets across proprietary backbones and distilled Qwen3 students. Qwen3-4B/8B-600 and Qwen3-4B/8B-6000 denote students fine-tuned on GPT\mbox{-}4.1 trajectories collected from 600 or 6000 training queries per graph, respectively.}
        \label{tab:full-stark-results}
    \end{strip}%
}

\theoremstyle{plain}

\theoremstyle{definition}

\theoremstyle{remark}

\title{
Autonomous Knowledge Graph Exploration \\ with Adaptive Breadth-Depth Retrieval
}
\author{
\textbf{Joaqu\'in~Polonuer\textsuperscript{1,2,$\ast$}}~\orcidlink{0009-0007-8613-6126}, 
\textbf{Lucas~Vittor\textsuperscript{1,$\ast$}}~\orcidlink{0009-0002-6978-0482},
\textbf{I\~naki~Arango\textsuperscript{1,$\ast$}}~\orcidlink{0009-0002-1443-2325},
\textbf{Ayush~Noori\textsuperscript{1,3,$\ast$}}~\orcidlink{0000-0003-1420-1236},\\
\textbf{David~A.~Clifton\textsuperscript{3,4}}~\orcidlink{0000-0002-9848-8555},
\textbf{Luciano~Del~Corro\textsuperscript{5,6,$\dagger$,$\ddagger$}},
\textbf{Marinka~Zitnik\textsuperscript{1,6,7,8,$\dagger$,$\ddagger$}}~\orcidlink{0000-0001-8530-7228} \\[1mm]
\textsuperscript{1}Department of Biomedical Informatics, Harvard Medical School, Boston, MA, USA \\
\textsuperscript{2}Departamento de Computaci\'on, FCEyN, Universidad de Buenos Aires, Buenos Aires, Argentina \\
\textsuperscript{3}Department of Engineering Science, University of Oxford, Oxford, UK \\
\textsuperscript{4}Oxford Suzhou Centre for Advanced Research, University of Oxford, Suzhou, Jiangsu, China \\
\textsuperscript{5}ELIAS Lab, Departamento de Ingenier\'ia, Universidad de San Andr\'es, Victoria, Argentina \\
\textsuperscript{6}Kempner Institute for the Study of Natural and Artificial Intelligence, Allston, MA, USA \\
\textsuperscript{7}Broad Institute of MIT and Harvard, Cambridge, MA, USA \\
\textsuperscript{8}Harvard Data Science Initiative, Cambridge, MA, USA \\
\textsuperscript{$\ast$}Equal contribution\; \textsuperscript{$\dagger$}Co-senior authors\\
\textsuperscript{$\ddagger$}Correspondence: \href{mailto:delcorrol@udesa.edu.ar}{delcorrol@udesa.edu.ar}, \href{mailto:marinka@hms.harvard.edu}{marinka@hms.harvard.edu}
}

\begin{document}
\maketitle

%!TEX root = ../main.tex

\begin{abstract}
Retrieving evidence for language model queries from knowledge graphs (KGs) requires balancing broad search across the graph with multi-hop traversal to follow relational links. Similarity-based retrievers provide coverage but remain shallow, whereas traversal-based methods rely on selecting seed nodes to start exploration, which can fail when queries span multiple entities and relations. We introduce \textbf{\retrieverShort}: \retriever, a tool-using KG retriever that gives a language model control over this breadth-depth tradeoff using a two-operation toolset: global lexical search over node descriptors and one-hop neighborhood exploration that composes into multi-hop traversal. \retrieverShort alternates between breadth-oriented discovery and depth-oriented expansion without depending on a fragile seed selection, a pre-set hop depth, or requiring retrieval training. \retrieverShort adapts tool use to queries, using global search for language-heavy queries and neighborhood exploration for relation-heavy queries.
On STaRK, \retrieverShort reaches 59.1\% average Hit@1 and 67.4 average MRR, improving average Hit@1 by up to 31.4\% and average MRR by up to 28.0\% over retrieval-based and agent-based training-free methods.
Finally, we distill \retrieverShort's tool-use trajectories from a large teacher into an 8B model via label-free imitation, improving Hit@1 by +7.0, +26.6, and +13.5 absolute points over the base 8B model on AMAZON, MAG, and PRIME datasets, respectively, while retaining up to 98.5\% of the teacher's Hit@1 rate.
\end{abstract}

\FigMain
%!TEX root = ../main.tex

\section{Introduction}
Large language models rely on knowledge retrieval to ground and align their outputs in external evidence~\cite{ren2025investigating,wang2025survey,xia2025ground,zhang2024knowledge}, from retrieval-augmented generation (RAG) to systems and memory modules that operate over \emph{semi-structured} knowledge bases (SKB) that mix text with relational information~\cite{lewis_retrieval-augmented_2020,guu_realm_2020,karpukhin_dense_2020,izacard_leveraging_2021,mavromatis_gnn-rag_2025,chen_gril_2025,li2025retrollm}. Knowledge graphs (KGs) are a natural data representation for this setting because they organize evidence around entities and typed edges, support reuse across queries, and enforce relational constraints that a flat text index cannot express. This has motivated graph-aware retrievers and graph-grounded generation methods, including graph-based RAG and SKB retrievers that combine text and relational data~\cite{edge_local_2025,zhu_knowledge_2025,xia_knowledge-aware_2025, yao2025expandr,jeong2025database}.

Retrieving evidence from KGs is challenging because it requires coordinating two competing search modes~\cite{wu_stark_2024,lee_hybgrag_2025,zhu-etal-2025-mitigating,yan2026explore}. Many queries require \emph{breadth}: they mention multiple entities or loosely connected concepts, so the retriever must cover the graph broadly to reach the right region. Other queries require \emph{depth}: the supporting evidence only appears after following specific multi-hop relational paths. Similarity-based retrievers provide global coverage but often remain shallow and underuse relational structure, whereas traversal-based methods can be brittle because they must choose seed entities to start exploration; when seeds are incomplete or ambiguous, the search stays local and misses evidence elsewhere.

Existing work tackles these breadth-depth requirements in isolation rather than jointly~\cite{ko2025cooprags,ma2025think,wang2025corag}. \emph{Structure-aware} retrievers extend text-based retrieval with relational structure, for example, by learning node embeddings that aggregate information from nearby neighbors or by generating candidates using a local graph neighborhood before ranking them~\cite{lee_hybgrag_2025,zhu_knowledge_2025,lei_mixture_2025}. These methods capture local structure, but they typically encode a fixed neighborhood around each node, so deeper multi-hop queries require expanding the encoded context or stacking additional message-passing and retrieval stages, which increases complexity and cost~\cite{wan2025digest,hu2025remindrag,verma2024plan}. By contrast, \emph{traversal-based approaches} perform multi-hop exploration, but they depend on identifying a small set of seed entities from which exploration begins~\cite{markowitz_tree--traversals_2024,sun_think--graph_2024}. When seeds are incomplete or ambiguous, exploration stays local and misses relevant information elsewhere in the graph~\cite{liu2025ontology,ma2025think}. Many systems also rely on task- or graph-specific training to learn traversal or scoring policies, which limits transfer across domains and graph schemas~\cite{li_multi-field_2025,lei_mixture_2025,wu_avatar_2024,yu_can_2025}. As a result, existing methods struggle to combine global search with targeted relational reasoning for adaptive retrieval.

%~\cite{yao_react_2023,schick_toolformer_2023}

\textbf{Present work.}
We introduce \textbf{\retrieverShort}: \retriever, a tool-using KG retriever that gives a language model control over evidence discovery using a minimal set of tools for global lexical search and neighborhood exploration. \retrieverShort does not require selecting seed entities to start exploration or establishing a maximum hop depth in advance; instead, the model alternates between broad global search and targeted multi-hop traversal based on the query and what it has retrieved so far. We evaluate \retrieverShort on the heterogeneous graphs in the STaRK retrieval benchmark and show consistent gains across all settings. We further study compute-accuracy tradeoffs by varying the tool-call budget and the number of parallel agents, and we distill \retrieverShort's tool-use policy into an 8B model via label-free trajectory imitation, preserving most of the performance of teacher model at substantially lower inference cost~\cite{kang_distilling_2025}.

Our contributions are threefold. (i) We introduce \retriever, a training-free retrieval framework that equips language models with a minimal but expressive tool interface for adaptive retrieval from KGs. (ii) We show that \retrieverShort balances breadth-oriented retrieval with depth-oriented multi-hop traversal without task- or graph-specific training, achieving strong performance on STaRK. (iii) We distill \retrieverShort's tool-use policy without labeled supervision into a compact Qwen3-8B model~\cite{yang_qwen3_2025,kang_distilling_2025}, preserving retrieval quality while reducing inference cost.

%!TEX root = ../main.tex

\section{Related Work}
\label{sec:related}

\slimparagraph{Knowledge graphs for document-centric RAG.}
Retrieval-augmented generation grounds LLM outputs in external evidence \cite{lewis_retrieval-augmented_2020,guu_realm_2020}. Recent work injects structure by building graphs to aggregate evidence. GraphRAG performs local-to-global retrieval over entity-centric graphs \cite{edge_local_2025}, and related methods retrieve textual subgraphs for multi-hop queries \cite{zhu_knowledge_2025,hu_grag_2025,he_g-retriever_2024,li_simple_2025}. These approaches improve how evidence is organized but the retrieval process itself remains static.

\slimparagraph{Retrieval over semi-structured knowledge bases.}
Complementary to document-centric graph indices, semi-structured knowledge base retrievers directly combine text and explicit relations. KAR grounds query expansion in KG structure \cite{xia_knowledge-aware_2025}, GraphSearch mixes graph and text channels with iterative refinement \cite{yang_graphsearch_2025} and HybGRAG routes queries across a bank of retrieval modules with self-reflection \cite{lee_hybgrag_2025}; CoRAG highlights cooperative hybrid retrieval that preserves global semantic access beyond local neighborhoods \cite{zheng_corag_2025}. In parallel, parametric retrievers such as MoR and mFAR learn to fuse lexical, semantic, and structural signals for ranking \cite{lei_mixture_2025,li_multi-field_2025}. Across these variants, retrieval is framed as scoring candidates from a static index. \retrieverShort differs in that retrieval is formulated as an interactive process: the model dynamically switches between global search and neighborhood expansion, guided by the query requirements and without relying on task-specific supervision.

\slimparagraph{Agents for multi-hop KG retrieval.}
A separate line of work treats the KG as an environment for iterative interaction. Earlier agents follow relation paths using reinforcement learning or learned policies \cite{das_go_2018, xiong_deeppath_2017, sun_pullnet_2019, asai_learning_2020}. In the LLM era, tool-use frameworks such as ReAct \cite{yao_react_2023} and prompt-optimization methods such as AvaTaR \cite{wu_avatar_2024} enable interactive evidence gathering.
Within KG retrieval, traversal-based approaches expand from seed entities using prompted heuristics or learned policies, including Tree-of-Traversals, Think-on-Graph, and GraphFlow \cite{markowitz_tree--traversals_2024,sun_think--graph_2024,yu_can_2025, ma2025think}; related KG-grounded reasoning methods also emphasize multi-step planning or navigation on the KG \cite{luo_reasoning_2024,sun_search--graph_2025}. Traversal-based agents are effective when the correct starting entities are known, but they are prone to anchoring errors and can over-commit to local neighborhoods once exploration begins. In \retrieverShort, global search remains available throughout the trajectory, allowing the agent to retain a complete view of the KG at each step. This design enables coordination between global discovery and multi-hop expansion within a single retrieval. Recent work also explores using reinforcement learning to enable LLMs to explore reasoning paths over KGs~\cite{yan2026explore,wang2026kg}.

Existing work varies in whether it treats retrieval as static ranking over an index or as sequential decision-making on the graph, and in whether it requires graph-specific supervision to learn a ranking function or a traversal policy. \retrieverShort adapts its search strategy online through a minimal, graph-native tool interface. It is training-free; when needed, its tool-use policy can be distilled into compact models from interaction trajectories without ground-truth relevance labels, improving efficiency while preserving retrieval quality.

%!TEX root = ../main.tex
\section{\retrieverNoSc}
\label{method}

We study retrieval over a knowledge graph \unbreakable{$G=\langle V,E,\phi_V,\phi_E,d_V\rangle$}, where $V$ and $E$ denote entities and edges, $\phi_V$ and $\phi_E$ assign a type to each node and relation, and $d_V(v)$ denotes the text attributes associated with node $v$, such as titles, descriptions, or other metadata. Given a natural-language query $Q$, retrieval is formulated as an interactive process in which an agent $\mathcal{A}=\langle \mathrm{LLM},\mathcal{T}\rangle$ queries the graph through a tool interface $\mathcal{T}$ \cite{yao_react_2023,schick_toolformer_2023} and produces a trajectory $\tau = \bigl((s_1,A_1,o_1),\dots,(s_T,A_T,o_T)\bigr)$. At step $t$, $s_t$ contains $Q$ and the interaction history, $A_t$ is a sequence of tool invocations, and $o_t$ is the observation returned after executing $A_t$.

Throughout the trajectory, the agent maintains an ordered list of retrieved nodes $\mathcal{R}$. At each step, it can \textsc{select} nodes returned by tools and append them to $\mathcal{R}$, or terminate by issuing a dedicated \textsc{finish} action. Execution ends either when the agent calls \textsc{finish} or when the maximum trajectory length $T_{\max}$ is reached. The final retrieval output is the ranked list $\mathcal{R}=(v_1,v_2,\dots)$, where earlier selections receive higher rank.

To rank candidate nodes returned by tools, we use a relevance function $\rel(q,d_V(v))\in\mathbb{R}_{\ge 0}$ that scores node $v$ for a textual subquery $q$ provided by the agent as a tool parameter. We implement $\rel$ with BM25 \cite{robertson_probabilistic_2009} over an inverted index of node textual attributes \cite{manning_introduction_2008}, yielding fast and stable scoring for the many short, evolving subqueries issued during exploration.

\subsection{Tools}
We implement the interaction described above through a small set of retrieval tools. Each tool returns a candidate set of nodes, which the agent may append to $\mathcal{R}$ or use to guide subsequent steps. \\

\slimparagraph{Global search} retrieves the $k$ highest-scoring nodes in the graph under $\rel$ for an agent-issued subquery $q$, as shown in \Cref{fig:main-figure}a:
\begin{equation*}
    \operatorname{Search}_G(q,k)
    \;:=\;
    \operatorname*{Top\text{-}k}_{v \in V}\; \rel(q, d_V(v))
\end{equation*}
This tool provides broad entry points into the graph and is primarily used (i) to locate entities related to the user query $Q$, which will then be used for further exploration, and (ii) to handle cases where direct text matching suffices without requiring multi-hop reasoning. \\

\slimparagraph{Neighborhood exploration} returns adjacent nodes of a node $v$ filtered by optional node and edge type constraints $F := (F_V, F_E)$ selected by the agent as tool parameters, and optionally ranked using an agent-generated subquery $q$ (\Cref{fig:main-figure}b).

The filtered one-hop neighborhood $N_F$ of a node $v$ is defined as:
\begin{equation*}
    N_F(v)
    :=
    \left\{
    u \in N(v)
    \;\left|\;
    \begin{aligned}
    & \phi_V(u) \in F_V, \\
    & \phi_E(\{u, v\}) \in F_E
    \end{aligned}
    \right.
    \right\}
\end{equation*}

where $N(v)$ denotes the open neighborhood of $v$ and $\{u, v\}$ denotes the edge connecting $v$ and $u$, regardless of direction. Edge directionality and relation types are exposed in the tool output. To control the size of the retrieved neighborhood, we introduce a fixed retrieval budget $k \in \mathbb{N}$:
\begin{equation*}
    \operatorname{Neighbors}(v, q, F)
    := \operatorname*{Top\text{-}k}_{u \in N_F(v)} \; \rel(q, d_V(u))
\end{equation*}
Each call to $\operatorname{Neighbors}$ retrieves a single-hop neighborhood. The agent performs multi-hop retrieval by calling this operator multiple times in sequence, optionally interleaved with global re-anchoring (\Cref{fig:main-figure}c). The number of hops is not fixed in advance; the agent decides when to stop expanding based on what it has retrieved so far.

\subsection{Parallel Exploration}
\label{sec:parallel_exploration}
We increase robustness by running $n$ independent instances of the same agent in parallel and aggregating their retrieved lists, akin to self-consistency and voting-based ensembling in LLM reasoning \cite{wang_self-consistency_2023,kaesberg_voting_2025}. Each agent produces an ordered list of retrieved nodes $\mathcal{R}^{(i)}=(v^{(i)}_1,v^{(i)}_2,\dots)$ from an independent trajectory. We then combine these lists using a simple rank-fusion rule inspired by classical rank aggregation and data fusion methods \cite{fagin_efficient_2003,cormack_reciprocal_2009}.

Concretely, we concatenate the per-agent lists in agent order to form:
\begin{equation*}
    L \;:=\; \mathcal{R}^{(1)} \,\Vert\, \mathcal{R}^{(2)} \,\Vert\, \cdots \,\Vert\, \mathcal{R}^{(n)},
\end{equation*}
and let $\mathcal{V}_L$ be the set of unique nodes in $L$. The final ranking $\mathcal{R}$ orders nodes by decreasing frequency in $L$ (vote count), breaking ties by the earliest position at which a node appears in any trajectory, favoring nodes discovered earlier during exploration.

\subsection{Agent Distillation}
\label{subsec:distill}
While \retrieverShort operates on off-the-shelf models, its behavior can be distilled into a smaller language model to reduce inference cost and latency \cite{hinton_distilling_2015}. We adopt a standard teacher--student paradigm in which a student model imitates the tool-usage trajectories of a stronger teacher LLM via supervised fine-tuning \cite{schick_toolformer_2023}.

\slimparagraph{Trajectory generation.}
For each training query $Q$ on a given graph $G$, we run the teacher agent to collect trajectories $\tau$ as defined in \Cref{method}. Each trajectory contains the full interaction record: the agent's tool calls and parameters interleaved with the resulting tool observations.

\slimparagraph{Training objective.}
The student is trained with next-token prediction on the collected trajectories \cite{ouyang_training_2022}. We compute loss only on assistant-authored tokens; user messages and tool outputs are masked \cite{huerta-enochian_instruction_2024,shi_instruction_2024}. This trains the student to reproduce the teacher's decisions, which tools to invoke and how to parameterize them, while tool execution remains external to the model. 

\slimparagraph{Label-free supervision.}
Importantly, distillation does not require ground-truth evidence nodes for queries. Supervision is derived solely from teacher trajectories, making the approach applicable in realistic settings where relevance labels are unavailable: one can run a strong teacher to generate trajectories on a target graph and then fine-tune a smaller model directly from these interactions \cite{schick_toolformer_2023,kang_distilling_2025}.
\TabStarkResults
%!TEX root = ../main.tex

\section{Experimental Setup}
\label{sec:experiments}
We measure retrieval performance on STaRK, a benchmark for entity-level retrieval over heterogeneous, text-rich KGs \cite{wu_stark_2024}.

\subsection{Benchmark}

STaRK comprises three large, heterogeneous knowledge graphs. \textbf{AMAZON} is an e-commerce graph with roughly 1M entities and 9.4M relations, constructed from Amazon metadata, reviews \cite{he_ups_2016}, and question--answer pairs \cite{mcauley_inferring_2015}. \textbf{MAG} is a scholarly graph with 1.9M entities and 39.8M relations derived from the Microsoft Academic Graph \cite{wang_microsoft_2020}. \textbf{PRIME} is a biomedical graph built from PrimeKG \cite{chandak_building_2023}, containing 129K entities and 8.1M relations. Each node is associated with text-rich attributes, making STaRK a natural testbed for hybrid retrieval over structured and textual signals. Given a query, the retriever must return a ranked list of nodes that support the answer. We report agent configuration and hyperparameters in Appendix~\ref{app:impl-details}.

\subsection{Baselines and metrics}

We compare with representative retrieval-based and agent-based baselines, focusing on methods that report results on all three graphs, as our goal is to evaluate performance consistently across different regimes and assess generality.

\slimparagraph{Retrieval-based.} \textbf{BM25} \cite{robertson_probabilistic_2009} is the same sparse, lexical retriever used for global search. We also include dense embedding retrievers that rank nodes by cosine similarity, using \textbf{ada-002} and \textbf{GritLM-7B}, an instruction-tuned 7B encoder \cite{muennighoff_generative_2025}. \textbf{mFAR}~\cite{li_multi-field_2025} is a multi-field adaptive retriever that combines keyword matching with embedding similarity to learn query-dependent weights over different node fields. \textbf{KAR}~\cite{xia_knowledge-aware_2025} augments queries with knowledge-aware expansions and applies relation-type constraints during retrieval. \textbf{MoR}~\cite{lei_mixture_2025} is a trained retriever that combines multiple retrieval objectives.

\slimparagraph{Agent-based.} \textbf{Think-on-Graph} \cite{sun_think--graph_2024} is a training-free LLM agent that iteratively expands paths in the graph using beam search. \textbf{GraphFlow} \cite{yu_can_2025} learns a policy for generating multi-hop retrieval trajectories using GFlowNets \cite{bengio_flow_2021}. \textbf{AvaTaR} \cite{wu_avatar_2024} is a tool-using agent that optimizes prompting from positive and negative trajectories.

Results for KAR, mFAR, MoR, AvaTaR, and GraphFlow are reported as in their respective papers, which evaluate on the official STaRK splits and metrics. For Think-on-Graph, we report the numbers provided in the GraphFlow study, which includes a direct comparison to Think-on-Graph under the same STaRK setup \cite{yu_can_2025,sun_think--graph_2024}.

\slimparagraph{Metrics.} We follow the STaRK protocol and report Hit@1, Hit@5, Recall@20 (R@20), and Mean Reciprocal Rank (MRR), which capture top-rank precision, coverage of the ground-truth set, and overall ranking quality. Note that Hit@5 is reported in \cref{tab:full-stark-results} in the Appendix.

\slimparagraph{Backbones.} Our primary configuration uses GPT-4.1 as the backbone LLM throughout the paper. Appendix \Cref{tab:full-stark-results} additionally reports results with GPT-4o for comparability with prior KG retrievers such as KAR and Think-on-Graph, and with GPT-5.2 to verify that the retrieval interface transfers to a newer backbone on the same splits.

\subsection{Distillation Setup}
\label{sec:distillation-setup}

For each graph, we collect teacher trajectories on the training split to distill \retrieverShort into a smaller, lower-cost model (\Cref{subsec:distill}), offering a viable alternative when under tighter compute budgets. Using GPT-4.1 as the teacher, we run \retrieverShort three times per training query with stochastic decoding (temperature $= 0.7$), producing three trajectories per query. We cap the distillation budget by subsampling up to 6{,}000 training queries per graph, yielding at most 18{,}000 trajectories per graph (full statistics in \cref{tab:traj_counts}) and summing to 94.4 million tokens. Each trajectory is limited to $T_{\max}{=}20$ steps and ends when the agent issues \textsc{finish} or reaches the step limit. We apply no trajectory filtering or rejection sampling, preserving a label-free setting. We then distill a Qwen3-8B \cite{yang_qwen3_2025} student via supervised fine-tuning with LoRA adapters \cite{hu_lora_2021}, using next-token prediction over assistant-authored tokens only. We train for one epoch with a 16{,}384-token context length using AdamW \cite{loshchilov_decoupled_2019} at learning rate $1\times 10^{-5}$, selecting checkpoints via early stopping on the official validation split. Training runs on a single NVIDIA H100 GPU and completes in approximately five hours.

%!TEX root = ../main.tex

\section{Results}

\subsection{Benchmarking}
\FigNAgents
\Cref{tab:stark-results} reports retrieval performance on STaRK, grouped by training regime. Across all methods assessed, \retrieverShort achieves the best average performance. 

Classical retrievers remain strong baselines on AMAZON, when queries are predominantly descriptive. By incorporating local structural cues, KAR improves over lexical methods, but its shallow neighborhood expansion is limited on multi-hop queries \cite{xia_knowledge-aware_2025}.

Think-on-Graph and GraphFlow highlight the benefits of multi-hop traversal, performing well on PRIME. Think-on-Graph is appealing due to its training-free setup, and GraphFlow shows that strong performance can be achieved with smaller backbones through reinforcement learning. However, both degrade on AMAZON's text-heavy, broad queries, as they lack global search primitives and can be sensitive to brittle anchoring and entity identification \cite{sun_think--graph_2024, yu_can_2025}.

\retrieverShort performs consistently across regimes and is especially strong on MAG. This pattern aligns with its tool design. Global search offers a reliable anchor for text-heavy queries and supports strong top-rank accuracy on AMAZON. Typed, query-ranked one-hop expansion enables controlled multi-hop evidence gathering in relational settings, contributing to the best results on MAG and solid performance on PRIME, where it is surpassed only by the RL-trained GraphFlow.

While \retrieverShort uses a large backbone, the distilled variant preserves most of these gains with a substantially smaller Qwen3-8B model via label-free trajectory imitation (Section~\ref{sec:distillation-impact}).

\subsection{Text vs. Relational Adaptive Retrieval}
\label{sec:adaptive-tool-use}
\FigAdaptiveToolUse[h]

STaRK reports, for each graph, the average share of queries that are primarily textual versus primarily relational (multi-hop) (Fig.~5 in \citet{wu_stark_2024}). We use these proportions as a reference and compare them to \retrieverShort's tool-call allocation on our evaluation split. Concretely, we treat the fraction of global search calls as a proxy for text-centric evidence use and the fraction of neighborhood exploration calls as a proxy for relation-centric evidence. \Cref{fig:adaptive-tool-use} shows a proportional match: on AMAZON, where queries are mostly textual, \retrieverShort relies almost entirely on global search (87.7\%), whereas on MAG and PRIME, where relational requirements dominate, \retrieverShort shifts toward neighborhood exploration to traverse multi-hop evidence (65.3\% and 52.3\%, respectively). Appendix~\ref{app:trajectory-tool-usage} complements this call-frequency view with trajectory-level statistics on re-anchoring after the first step and on tool mixing among successful runs.

\subsection{Impact of Toolset Design}
\label{sec:toolset}
\FigSFTImpact
\TabToolsetAblation[h]

We conduct various ablation studies to assess the impact of toolset design choices. \Cref{tab:toolset-ablation} demonstrates that neighborhood exploration is the main source of gains on relational, multi-hop queries. Removing this tool decreases performance on MAG and PRIME as the system is then limited to global lexical search without graph traversal. On AMAZON, performance drops more moderately and moves toward lexical baselines (\cref{tab:stark-results}).

We further separate two complementary controls in neighborhood exploration. Disabling query-based ranking causes smaller but consistent drops, suggesting that lexical matching within a local neighborhood helps surface relevant neighbors and prevents drift toward high-degree distractors. Disabling type-based filtering is more detrimental, especially in heterogeneous graphs such as PRIME (\cref{tab:stark-statistics}). In such environments, type constraints are important to direct the agent towards semantically relevant edges and nodes, preventing search from drifting into unrelated parts of the graph.

Note that we do not ablate global search because it is required: it maps query text to candidate nodes and provides the node identifiers needed to start neighborhood expansion. Without this initial anchor, the agent cannot reliably enter the right part of the graph, so the system fails outright.

\subsection{Compute-Performance Trade-offs}
\label{sec:cost-performance}

We next study how retrieval quality scales with the inference-time budget. \Cref{fig:n-agents} shows that performance improves monotonically as compute increases, moving from single-agent settings to parallel multi-agent configurations. We also report a single-agent vs.~multi-agent comparison under the same total step budget in Appendix~\ref{app:budget-control}. 

Additional compute helps most on queries that require multi-hop expansion, and yields smaller gains when global lexical search is already sufficient.  Parallelization yields performance benefits with minimal overhead. Increasing the agent count, particularly the transition from one to two agents, results in substantial gains while only modestly increasing latency. Because agents run independently, end-to-end latency is determined by the bottleneck of the slowest agent rather than the cumulative runtime of all agents.

\retrieverShort provides an interpretable budget-performance landscape. Practitioners can fix a latency or compute budget and choose an operating point in \Cref{fig:n-agents} that matches their needs, trading off depth and parallelism to balance quality and cost across graph regimes. 

\subsection{Aggregation Strategy Comparison}
\label{sec:aggregation-strategy}

\TabAggregationComparison[h]

Aggregation quality also matters once multiple trajectories have been collected. \Cref{tab:aggregation-comparison} compares our voting rule against two simpler ways of combining the outputs of three agents. Voting consistently outperforms both ordering-based merging and random merging, with the largest gains on MAG and PRIME. This supports the view that parallel exploration helps not only by increasing sampling diversity, but also by exposing a stable consensus signal over retrieved nodes. 

\subsection{Relevance Function (Lexical vs.\ Dense)}
\label{sec:relevance-lexical-dense}
\TabDenseRetrieverAblation[h]

The relevance function $\rel$ used to rank candidate nodes for agent-issued subqueries (\Cref{method}) can be implemented with either lexical (BM25) or dense retrieval. \Cref{tab:dense-ablation} compares the two options inside \retrieverShort. BM25 is notably more precise on AMAZON and PRIME, while differences on MAG are marginal in both directions. This suggests that the advantage of dense retrieval in single-pass settings does not directly carry over to iterative retrieval: the agent can compensate for weaker lexical matching by issuing refined subqueries over successive steps and by discovering relevant nodes through neighborhood exploration rather than text similarity alone.

\subsection{Impact of Distillation}
\label{sec:distillation-impact}

We also study how the distillation budget affects final performance. \Cref{fig:sft-impact} compares Qwen3-4B and Qwen3-8B students trained on increasing numbers of teacher trajectories across the three STaRK graphs; full results for all metrics are in \Cref{tab:full-stark-results}.

Distillation is data-efficient: using 10\% of the trajectories recovers roughly half of the total improvement achieved with the full training set. This makes distillation practical when collecting trajectories is costly. In our setup, distilling the 600-query setting can be done in 30 minutes on a single H100 GPU.

Student size matters most on PRIME. Because base models perform poorly in this regime, distillation is more important, and the teacher-student gap is the largest. The stronger performance of the 8B student suggests that higher-capacity models better absorb the long-horizon, high-branching exploration patterns required for complex biomedical reasoning in PRIME.

\subsection{Neighborhood Exploration vs. Retrieval}
\FigNeighborExploreDistribution[h]

We next examine how neighborhood exploration relates to retrieval success on MAG and PRIME. Here, successful trajectories refers to the runs (\textit{i.e.}, tool call sequences) that retrieve the correct nodes and therefore score as correct on the retrieval metric for that query. \Cref{fig:neighbor-explore-distribution} shows two failure modes. First, many failed runs make no neighborhood calls at all, suggesting the agent does not recognize when relational evidence is needed and never moves beyond global anchoring into multi-hop expansion. Second, failed runs also show a long tail with many neighborhood calls, indicating the opposite problem: the agent keeps expanding without converging on relevant support, consistent with drift in high-branching parts of the graph. 

In contrast, successful trajectories use neighborhood exploration sparingly, rarely exceeding ten calls, suggesting that strong retrieval relies on selective expansion rather than indiscriminate multi-hop traversal. These results underscore the need for retrieval methods like \retrieverShort that balance the breadth-depth tradeoff: when \retrieverShort succeeds, it adaptively switches from global anchoring to neighborhood expansion, and it stops expanding once it has found the needed support.

%!TEX root = ../main.tex

\section{Conclusion}
\label{conclusion}
We introduced \retriever, a training-free retrieval framework that exposes knowledge graphs through a minimal set of primitives for global search and local relational expansion. Across all three STaRK graphs, \retrieverShort achieves strong and stable retrieval performance while exhibiting a clear and interpretable inference-time budget–performance trade-off. We further show that this adaptive retrieval behavior can be transferred to a compact backbone via label-free trajectory distillation with modest data and compute, preserving nearly all of the teacher’s performance. Together, these results indicate that adaptive graph retrieval can be both practical and modular, and that exposing a small set of well-chosen retrieval operations is sufficient to unlock robust, general-purpose knowledge graph retrieval.
%!TEX root = ../main.tex

\section{Limitations}

Despite strong retrieval performance, \retrieverShort has limitations. First, agentic retrieval incurs higher latency than single-pass retrievers because it requires multiple LLM calls over an interaction trajectory. Larger budgets improve retrieval quality but also increase runtime. Second, our best-performing configuration relies on a large proprietary LLM, which can constrain scalability due to cost and availability. While \retrieverShort is LLM-agnostic, retrieval quality can drop with smaller models; we partially mitigate this via trajectory distillation into Qwen3-8B \cite{yang_qwen3_2025}, though distilled agents still trail the teacher on challenging regimes. Third, \retrieverShort assumes that node descriptors and relation information are sufficiently informative for BM25 global search and for ranking neighborhood expansions. Sparse or templated text can prevent the agent from locating relevant seed nodes or disambiguating them. Because the global search is lexical, mismatches in vocabulary (\textit{e.g.}, paraphrases and domain-specific aliases) can cause under-retrieval. Fourth, our evaluation is centered on text-rich KG benchmarks, so performance gains may not transfer to graphs with limited text descriptions. 

Although \retrieverShort is a general retrieval approach, agentic graph exploration can create risks if used without safeguards. Retrieval errors can be treated as support for downstream decisions, and interaction traces may expose sensitive attributes if node text contains private information. Mitigation requires redaction for sensitive fields and bias audits prior to deployment.

\section{Ethical Considerations}

This study does not use human annotators, crowdworkers, or research with human participants. Ethical concerns arise in downstream use of agentic retrieval over text-rich knowledge graphs. Retrieval errors can be treated as evidence and multi-step exploration can surface sensitive attributes present in graph text.
The approach may also amplify biases in the underlying graph. If some languages and communities have sparse descriptions or different naming conventions, global lexical search and neighborhood ranking may under-retrieve relevant information, leading to unequal coverage across groups and reduced benefits for underrepresented stakeholders.
We recommend safeguards before deployment, including redaction of sensitive fields and bias audits.
Potential positive impact includes improving access to large knowledge graphs for language models, including information that may be difficult to access with text retrieval alone.
\section{Code Availability}

All code used in this study is publicly available at \url{https://github.com/mims-harvard/ark}. A terminal-based chat implementation of \retrieverShort is at \url{https://github.com/mims-harvard/ark-agent-cli}.
\section*{Acknowledgments}

% \begin{spacing}{0.8}
% {\footnotesize 
We gratefully acknowledge the support of NSF CAREER 2339524, ARPA-H Biomedical Data Fabric (BDF) Toolbox Program, Harvard Data Science Initiative, Amazon Faculty Research, Google Research Scholar Program, AstraZeneca Research, Roche Alliance with Distinguished Scientists (ROADS) Program, Sanofi iDEA-iTECH Award, GlaxoSmithKline Award, Boehringer Ingelheim Award, Merck Award, Optum AI Research Collaboration Award, Pfizer Research, Gates Foundation (INV-079038), Aligning Science Across Parkinson's Initiative (ASAP), Chan Zuckerberg Initiative, John and Virginia Kaneb Fellowship at Harvard Medical School, Biswas Computational Biology Initiative in partnership with the Milken Institute, Harvard Medical School Dean’s Innovation Fund for the Use of Artificial Intelligence, and the Kempner Institute for the Study of Natural and Artificial Intelligence at Harvard University. 
A.N. was supported by the Rhodes Scholarship.
D.A.C. was funded by an NIHR Research Professorship (NIHR302440), a Royal Academy of Engineering Research Chair, and the InnoHK Hong Kong Centre for Cerebro-Cardiovascular Engineering, and was supported by the National Institute for Health Research Oxford Biomedical Research Centre and the Pandemic Sciences Institute at the University of Oxford.
Any opinions, findings, conclusions, or recommendations expressed in this material are those of the authors and do not necessarily reflect the views of the funders.
% }
% \end{spacing}

% ACL-style bibliography (uses acl_natbib.bst via the acl package)
\bibliography{custom, zotero}

\clearpage
\appendix
%!TEX root = ../main.tex
\TabAppendixResults
\renewcommand{\thefigure}{A\arabic{figure}}
\setcounter{figure}{0}

\section{Appendix}

\subsection{Dataset Statistics}

\begin{table}[H]
    \centering
    \begin{tabular}{lrrr}
        \toprule
        Dataset & Train   & Validation     & Test    \\
        \midrule
        PRIME   & 6{,}162 & 2{,}240 & 2{,}016 \\
        MAG     & 7{,}993 & 2{,}664 & 2{,}664 \\
        AMAZON  & 5{,}915 & 1{,}547 & 1{,}638 \\
        \bottomrule
    \end{tabular}
    \caption{Number of queries per dataset split for each STaRK graph.}
    \label{tab:traj_counts}
\end{table}

\newcommand{\TabStarkStatistics}[1][!ht]{%
    \begin{table}[#1]
        \centering
        \adjustbox{width=\columnwidth,center}{
            \begin{tabular}{lcccccc}
                \toprule
                Dataset 
                & \makecell{Entity\\types}
                & \makecell{Relation\\types}
                & \makecell{Average\\degree}
                & \makecell{Entities}
                & \makecell{Relations}
                & \makecell{Tokens} \\
                \midrule
                AMAZON & 4              & 5                & 18.2        & 1{,}035{,}542 & 9{,}443{,}802  & 592{,}067{,}882 \\
                MAG    & 4              & 4                & 43.5        & 1{,}872{,}968 & 39{,}802{,}116 & 212{,}602{,}571 \\
                PRIME  & 10             & 18               & 125.2       & 129{,}375     & 8{,}100{,}498  & 31{,}844{,}769  \\
                \bottomrule
            \end{tabular}
        }
        \caption{Statistics of the constructed semi-structured knowledge graphs used in STaRK.}
        \label{tab:stark-statistics}
    \end{table}%
}

\TabStarkStatistics

\subsection{Implementation Details}
\label{app:impl-details}

\retriever is run with $n{=}3$ parallel agents and a maximum trajectory length of $T_{\max}{=}20$. 

For the distilled variant, we use Qwen3-8B \cite{yang_qwen3_2025} (matching the model scale used by GraphFlow and Think-on-Graph with LLaMA3), trained via imitation on \retrieverShort teacher trajectories, while keeping the tool interface and exploration hyperparameters fixed.

Each tool call is executed with a bounded retrieval budget. Neighborhood exploration uses a fixed budget of $k{=}20$ neighbors per expansion. Global search returns up to $k$ nodes from the full graph. By default $k{=}5$, but the agent may override this value as a tool parameter.

Each agent outputs an ordered list of selected nodes. We aggregate these lists by ranking nodes first by the number of agents that selected them (vote count), and breaking ties by the earliest position at which the node appears in any agent’s list \cref{sec:parallel_exploration}. The aggregated ranking is truncated to the top 20 nodes to compute Recall@20; Hit@1 and MRR are computed on the same ranking.

\subsection{Additional Multi-Agent Analyses}
\label{app:budget-control}

\TabBudgetControlledComparison[h]

\noindent \Cref{tab:budget-controlled} compares a single longer trajectory with three shorter parallel trajectories under the same total step budget (30 steps). The three-agent configuration performs better on all three graphs, indicating that the gains from multi-agent inference are not explained solely by taking more total steps.

\subsection{Trajectory-Level Tool Usage}
\label{app:trajectory-tool-usage}

\TabTrajectoryToolMixing[h]

\noindent \Cref{tab:trajectory-tool-mixing} summarizes tool usage at the trajectory level. Because every trajectory begins with \textsc{Global Search}, runs either remain global-search-only or invoke both tools. On AMAZON, most trajectories, including successful ones, remain global-only, consistent with its text-heavy queries. On MAG and PRIME, by contrast, successful retrieval usually requires combining \textsc{Global Search} with neighborhood expansion.

\TabGlobalAfterFirstStep[h]

\noindent \Cref{tab:global-after-first-step} reports how often the agent returns to \textsc{Global Search} after the initial step. Re-anchoring is uncommon on AMAZON, but more frequent on MAG and especially PRIME, where retrieval more often requires shifting to a different region of the graph mid-trajectory. Together, \Cref{tab:trajectory-tool-mixing,tab:global-after-first-step} provide trajectory-level evidence for the two-tool design: AMAZON is often solved through lexical anchoring alone, whereas MAG and PRIME more often require combining global re-anchoring with local neighborhood expansion.

\subsection{Knowledge Graph Exploration Agent System Prompt}

The system prompt instructs the agent on the available node and relation types, the two tool signatures (\textsc{Global Search} and \textsc{Neighborhood Exploration}), output formatting, and the stopping criterion. The full prompt is included in the public repository of \retrieverShort at \url{https://github.com/mims-harvard/ark}.

\clearpage
%\FigArkAgentCli

\end{document}